\pdfoutput=1

\documentclass[11pt]{article}

\usepackage[]{emnlp2021}

\usepackage{times}
\usepackage{latexsym}

\usepackage[T1]{fontenc}

\usepackage[utf8]{inputenc}

\usepackage{microtype}

\usepackage{graphicx}
\usepackage{caption}
\usepackage{subcaption}
\usepackage{comment}
\usepackage{enumitem}
\usepackage{soul}
\usepackage{xcolor}

\title{Faithful Target Attribute Prediction in Neural Machine Translation}

\author{Xing Niu, Georgiana Dinu, Prashant Mathur, Anna Currey \\
  Amazon AI Translate \\
  \texttt{\{xingniu,gddinu,pramathu,ancurrey\}@amazon.com} \\}

\begin{document}
\maketitle
\begin{abstract}
The training data used in NMT is rarely controlled with respect to specific attributes, such as word casing or gender, which can cause errors in translations. We argue that predicting the target word and attributes \textit{simultaneously} is an effective way to ensure that translations are more faithful to the training data distribution with respect to these attributes. Experimental results on two tasks, uppercased input translation and gender prediction, show that this strategy helps mirror the training data distribution in testing. It also facilitates data augmentation on the task of uppercased input translation.
\end{abstract}

\section{Introduction}

Neural Machine Translation (NMT) \citep{bahdanau-etal-2015-neural,vaswani-etal-2017-attention} has become the leading MT technology, bringing discussions of human parity and even super-human performance \citep{hassan-etal-2018-achieving, barrault-etal-2019-findings}. Despite impressive advances in general translation quality, MT is not a solved problem: among other issues, NMT models make errors in \textit{attributes} of interest, such as making biased word gender prediction \citep{vanmassenhove-etal-2018-getting,stanovsky-etal-2019-evaluating,moryossef-etal-2019-filling} or generating incorrectly cased output given noisy or perturbed input \citep{berard-etal-2019-machine,niu-etal-2020-evaluating}.

The variation in MT performance along various dimensions of quality as defined by these attributes is rooted in the largely opportunistically sourced parallel data, where such attributes can be over- or under-represented. Manipulating the training data distribution is a popular way to address the problem, based on the assumption that a model predicts the same attribute distribution as that of its training data. Unfortunately, this is not guaranteed; training data skewness may even be amplified \citep{zhao-etal-2017-men,vanmassenhove-etal-2021-machine}.

We argue that effective training data manipulation requires model predictions to be \textit{faithful} to the distribution of attributes of interest. Faithfulness with respect to an attribute can be defined as preserving the training distribution of that attribute in inference predictions. We show that factored NMT -- simultaneously predicting the target word and attributes -- is effective for increasing faithfulness of translations to the training data distribution. To the best of our knowledge, this is the first work connecting factored NMT with models' faithfulness to the training data.

We study two token-level attributes -- casing and gender -- in two different tasks: uppercased text translation and gender prediction of professions. While modeling these two attributes with factored NMT has been shown to improve translation \textit{quality} \citep{koehn-hoang-2007-factored,sennrich-haddow-2016-linguistic,berard-etal-2019-machine,wilken-matusov-2019-novel,stafanovics-etal-2020-mitigating}, we focus on evaluating models' \textit{faithfulness} across various training data distributions, simulating the uncertainty of natural data. Results using simultaneous attribute prediction show that the translation output better mirrors the training distribution with respect to the attributes modeled: e.g., casing preservation ratio in inference matches the ratio in training.

We also demonstrate a novel and practical use case for leveraging a more faithful model: it enables easier development of data augmentation, a popular way to manipulate the training distribution along specific dimensions. In the task of uppercased text translation, we show that models with simultaneous attribute prediction and vocabulary space factoring are more robust to the size of augmented data (i.e., no unexpected quality degradation when augmenting with relatively little data).

\section{Factored NMT}

\citet{bilmes-kirchhoff-2003-factored} and \citet{alexandrescu-kirchhoff-2006-factored} introduced factored language models to deal with data sparsity and increase the generalization power of language models. They represent a word as a vector of factors $\{f_1,\dots,f_n\}$ (e.g., stem, word class) and explicitly model the probabilities of the vector sequences. In MT, factored models have been used to enrich phrase-based MT or NMT with linguistic features \citep{koehn-hoang-2007-factored,garcia-martinez-etal-2016-factored}. They reduce the output space by decomposing surface words $y$ on different dimensions, such as lemma and morphological tags, and maximize $\textrm{P}(y^t|\mathbf{x}) = \prod_{i=1}^n \textrm{P}(f_i^t|y^{<t},\mathbf{x})$. Factors are recombined to obtain the surface word. The term ``factor" has also been applied more generally to adding features to the source or target, as input or as auxiliary prediction functions even when the vocabulary space is not actually factored, i.e., $f_1$ is the surface word. \citep{koehn-hoang-2007-factored,sennrich-haddow-2016-linguistic}.

A similar line of work, target attribute prediction in NMT, falls under the paradigm of simultaneous prediction tasks~\citep{caruana-1997-multitask}. The additional training signal for predicting target attributes encourages the model to disentangle and leverage the attribute information more explicitly. Besides using factored NMT, this has also been done through inline concatenation of attributes to words \citep{nadejde-etal-2017-predicting,berard-etal-2019-machine}, which can achieve similar translation quality but increases the length of the generated sequence.

\section{Experiments}

We conduct experiments with two token-level attributes: uppercase style and gender of nouns. Capitalization conventions vary from language to language and often reflect content-specific style conventions, with all-caps commonly used for titles or to indicate tone of voice \citep{mcculloch2020because}. Although less represented in the data, translation models should be robust to such variations and correctly render both content and style of the source in the output. The gender attribute has been studied in the context of de-biasing MT models, where previous work has shown that models have difficulties correctly preserving gender: e.g., \textit{nurse} in \textit{He is a nurse} may be incorrectly translated as female in languages that require gender marking.

Section~\ref{sec:distribution} defines faithfulness in the context of these two attributes and investigates the effect of factors on faithfulness to the training distribution. In Section~\ref{sec:da} we investigate how factors impact data augmentation methods for improving MT on uppercased input.

We implement simultaneous target attribute prediction using factored NMT as exemplified in Table~\ref{tab:factors}. While different implementations of target factors exist \citep{garcia-martinez-etal-2016-factored,shi-etal-2020-case}, we opt for the Sockeye NMT toolkit \citep{domhan-etal-2020-sockeye}. It predicts target words $f_1$ and attributes $f_{2\dots n}$ with independent output layers, and the embeddings of the word and attributes are summed for the next decoder step. It incorporates the dependency between words and attributes by time-shifting attributes so that attributes at position $t$ are predicted at $t+1$.

\begin{table}[t]
\centering
\begin{tabular}{l|l|l|l}
    task & $y$ (surface) & $f_1$ (word) & $f_2$ (attribute) \\
    \hline
    case & NEURAL & neural & uppercase \\
    gender & chanteuse & chanteuse & feminine \\
\end{tabular}
\caption{Attributes are disentangled from the surface words. For case, words are lowercased to factor out casing information; for gender, $y$ and $f_1$ are identical.}
\label{tab:factors}
\end{table}

\subsection{Uncertain Attribute Distribution}
\label{sec:distribution}

\textbf{Uppercased Input Translation:} All-uppercased text is rare in many MT corpora, and the attribute of uppercasing is not always preserved from source to target. For example, ParaCrawl v7.1 \citep{banon-etal-2020-paracrawl} is a corpus crawled from various domains and is a representative corpus for building generic MT systems without controlling specific data attributes. In its \texttt{EN-DE} parallel set, only $2.1\%$ of English (EN) tokens and $1.7\%$ of German (DE) tokens are uppercased. Among 16K all-uppercased EN sentences, merely $1.8\%$ of their corresponding DE sentences preserve all-uppercasing (we denote this as \textit{uppercasing preservation ratio}, \textit{UPR}). While other languages contain the same magnitude of ratio of uppercased tokens, UPR varies across language pairs (e.g., $19\%$ for \texttt{DE-FR}).

We simulate the uncertainty of uppercasing preservation with controlled \texttt{EN-DE} ParaCrawl training data. First, we subsample 5M sentence pairs and lowercase all text.\footnote{Preliminary analyses show that subsampling the training data does not change observations, but it can speed up experimentation and alleviate the environmental impact of our computations of 72 models \citep{strubell-etal-2019-energy}.} Then, we select $2\%$ of pairs and uppercase the source sentences. This yields roughly $2\%$ uppercased source tokens, which approximates the true ratio of uppercased tokens in the raw data. Next, for these 2\% we vary UPR from $0\%$ to $100\%$ in increments of $20\%$. We compare four methods differing in the usage of source/target factors: (1) \texttt{none}, (2) only \texttt{source}, (3) only \texttt{target}, and (4) \texttt{both} source and target. Six models with the six different UPRs are trained for each method. We use raw WMT newstest19 for development (not optimized for uppercasing) and uppercased newstest20 \citep{barrault-etal-2020-findings} for testing. While we omit the exact results for brevity, none of the tested methods have a negative impact on translation quality on lowercased newstest20.

\begin{figure}[t]
	\centering
    	\begin{subfigure}[b]{\columnwidth}
        \centering
        \includegraphics[width=1\textwidth]{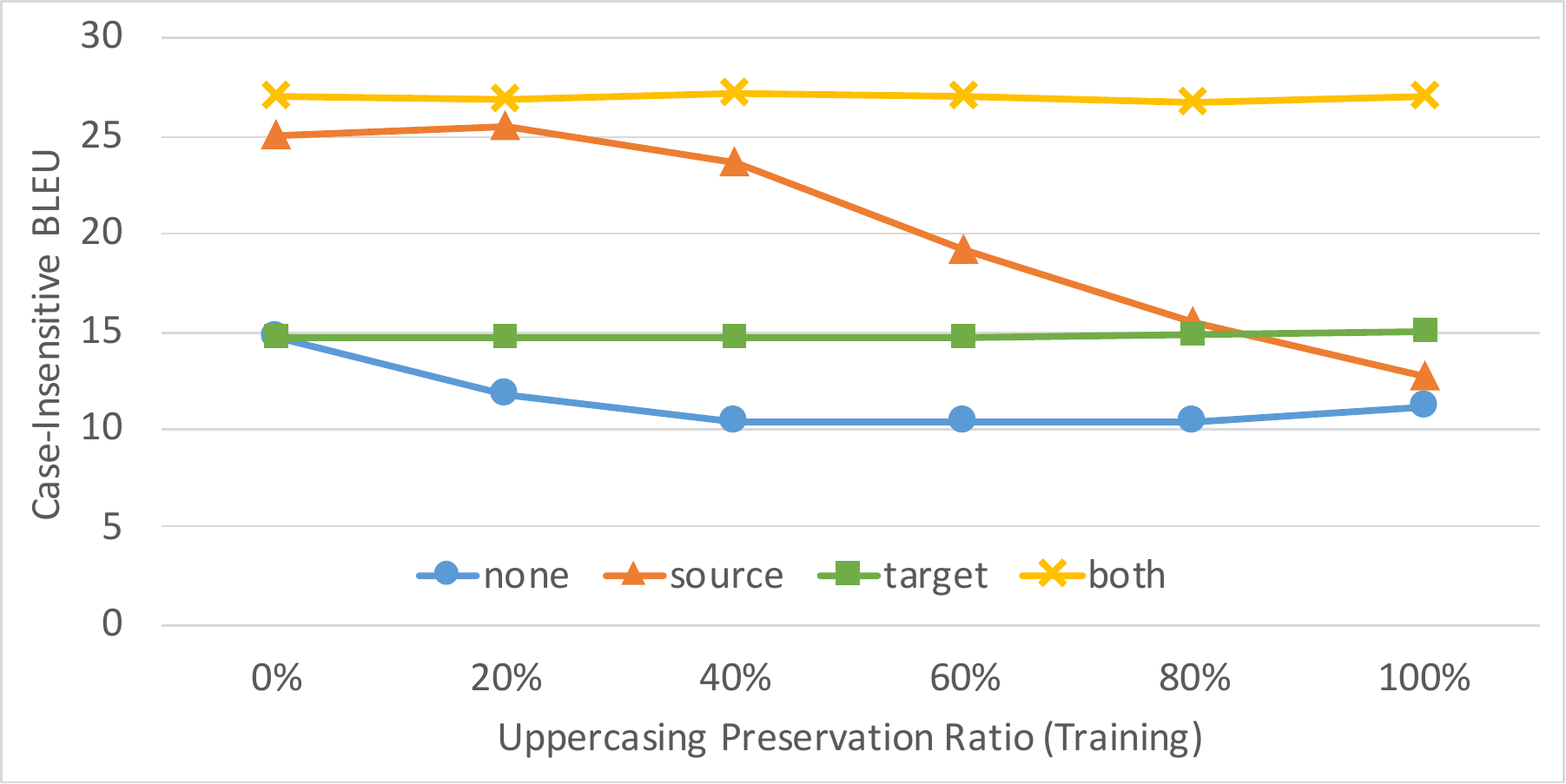}
        \caption{Using target factors (i.e., \texttt{both} and \texttt{target}) yields translations robust to the uncertainty of uppercasing preservation in terms of case-insensitive BLEU.}
        \label{fig:case-match-bleu}
        \end{subfigure}
    \hfill
        \begin{subfigure}[b]{\columnwidth}
        \centering
        \includegraphics[width=1\textwidth]{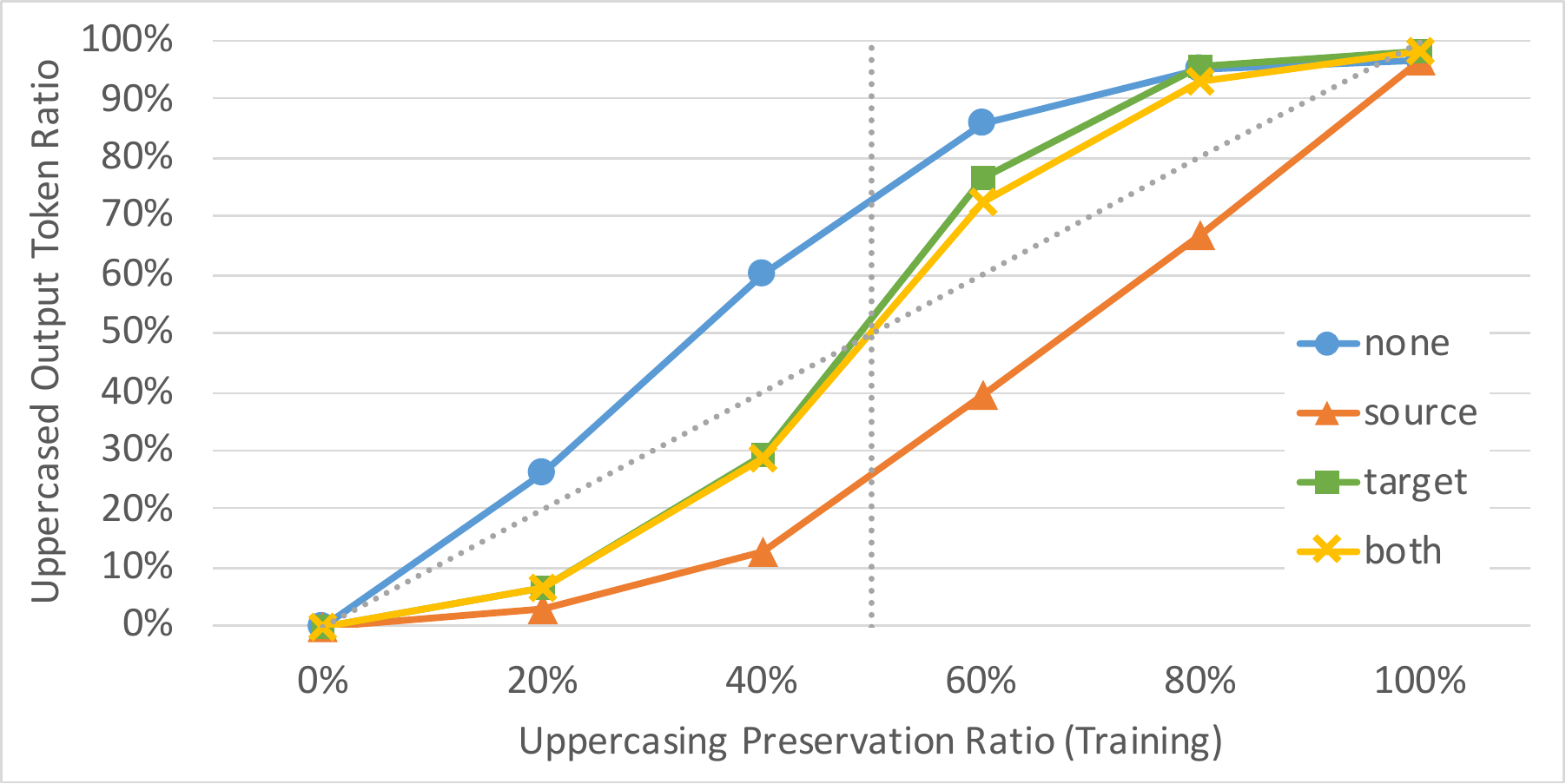}
        \caption{Using target factors outputs uppercased tokens with ratio more faithful to the training data distribution.}
        \label{fig:case-match-ratio}
        \end{subfigure}
	\caption{Uppercasing preservation ratio in the training data vs.\ case-insensitive BLEU and ratio of uppercased output tokens on uppercased newstest20.}
	\label{fig:case-match}
\end{figure}

For an effective evaluation, we disentangle casing from content and report both case-insensitive SacreBLEU \citep{post-2018-call} and uppercased output tokens ratio in Figure~\ref{fig:case-match}. The ratio of uppercased output tokens among all output tokens approximates UPR in prediction when the source test sentences are all-uppercased. Considering content translation measured by case-insensitive BLEU, when target factors are used (i.e., \texttt{both} and \texttt{target}), models are not sensitive to the uncertainty of UPR (Figure~\ref{fig:case-match-bleu}). More importantly, with simultaneous target attribute prediction, models predict uppercased tokens with closer probability to that of the training data (Figure~\ref{fig:case-match-ratio}).

\paragraph{Gender Prediction:} To further verify the hypothesis that simultaneously predicting the target attribute leads to more faithful attribute distribution, we also analyze gender information in words (i.e., gender markers) by investigating \texttt{EN-FR} models when translating professions grouped by their diverse natural occurrence in the training data.

\texttt{EN-FR} models are trained with 50M subsampled ParaCrawl sentence pairs. Gender markers for all target words are annotated using spaCy.\footnote{\url{https://spacy.io/}} All text is lowercased to exclude casing variations for easier analysis of gendered word distribution. We use lowercased WMT newstest13 for development and lowercased newstest14 for sanity check. The baseline model gets 36.6 BLEU while the model with target factors gets 36.5 BLEU, and the difference is not statistically significant.

To evaluate if the gender distribution of translated professions matches that of the training data, we create a test set as follows: we aggregate English occupations out of context from three lists collected by \citet{rudinger-etal-2018-gender}, \citet{zhao-etal-2018-gender}, and \citet{prates-etal-2020-assessing} and request both masculine and feminine French translations from paid professional translators, obtaining triples such as \{\textit{singer}, \textit{chanteur} (masc), \textit{chanteuse} (fem)\}. This yields a set of 353 gender-distinct translation pairs for which phrases in each pair appear at least five times in total in the training data.\footnote{Although we use a wide variety of professions, for 325 of the 353 professions, the masculine translation occurs more often than the feminine translation in our training data. We provide detailed description in Appendix \ref{appendix:professions} and release this test set at \url{https://github.com/amazon-research/gendered-profession-translations}.} We use our NMT models to score these paired translations via forced decoding \citep{sennrich-2017-grammatical} and pick the gender with a higher translation probability in a pair. 

\begin{figure}[t]
	\centering
	\includegraphics[width=1\columnwidth]{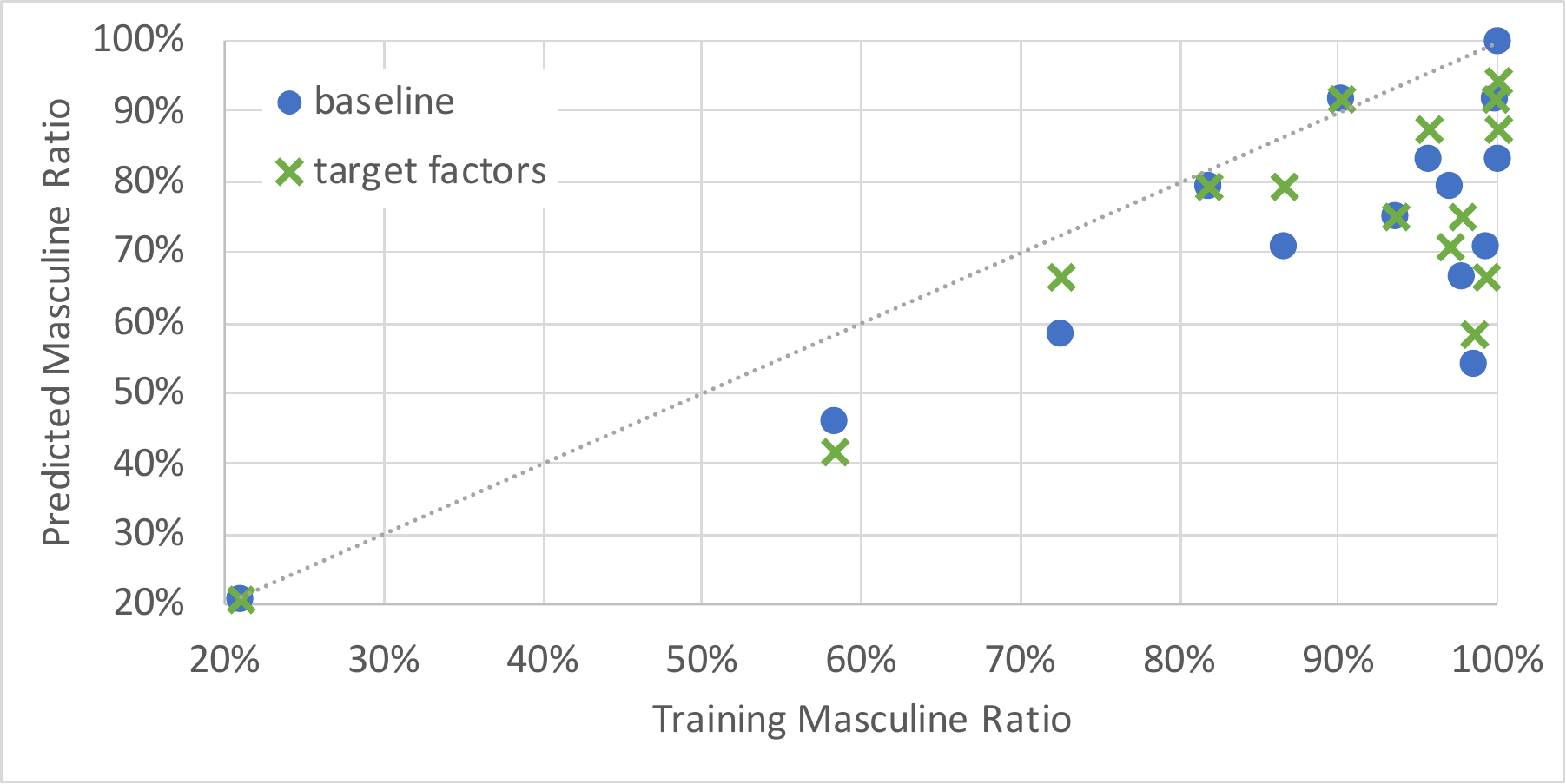}
	\caption{The predicted masculine ratio is closer to the training masculine ratio (MSE reduced 11\% from $0.037$) when using target factors vs.\ the baseline.}
	\label{fig:gender}
\end{figure}

Next, we sort French translation pairs by the \textit{training masculine ratio}, i.e.,
\[\frac{\mathrm{count}(masculine)}{\mathrm{count}(feminine)+\mathrm{count}(masculine)},\]
where $\mathrm{count}(masculine)$ is the count of the masculine French translation in the training data (and similarly for feminine). We evenly split these pairs into 15 bins. For each bin, we report the average training masculine ratio (on the $x$-axis) and the ratio of NMT models picking the masculine translation (on the $y$-axis) in Figure~\ref{fig:gender}. These data points approximate gender distribution for training vs.\ that of the model prediction. We find that using target factors brings the predicted masculine ratio closer to the training masculine ratio by reducing 11\% MSE from $0.037$.\footnote{We do not test source factors because English largely lacks grammatical gender distinctions. We leave improving in-context gender translation as future work: see preliminary results in Appendix \ref{appendix:gender}.}

\subsection{Data Augmentation (DA)}
\label{sec:da}

With uncertain attribute distribution in the training data, obtaining the desired distribution in the output (e.g., always preserving source case) will be uncertain as well. A practical solution is to manipulate the training distribution using data augmentation, for example by uppercasing a portion of the training data and adding it to the training.

\begin{figure}[t!]
	\centering
    	\begin{subfigure}[b]{\columnwidth}
        \centering
        \includegraphics[width=1\textwidth]{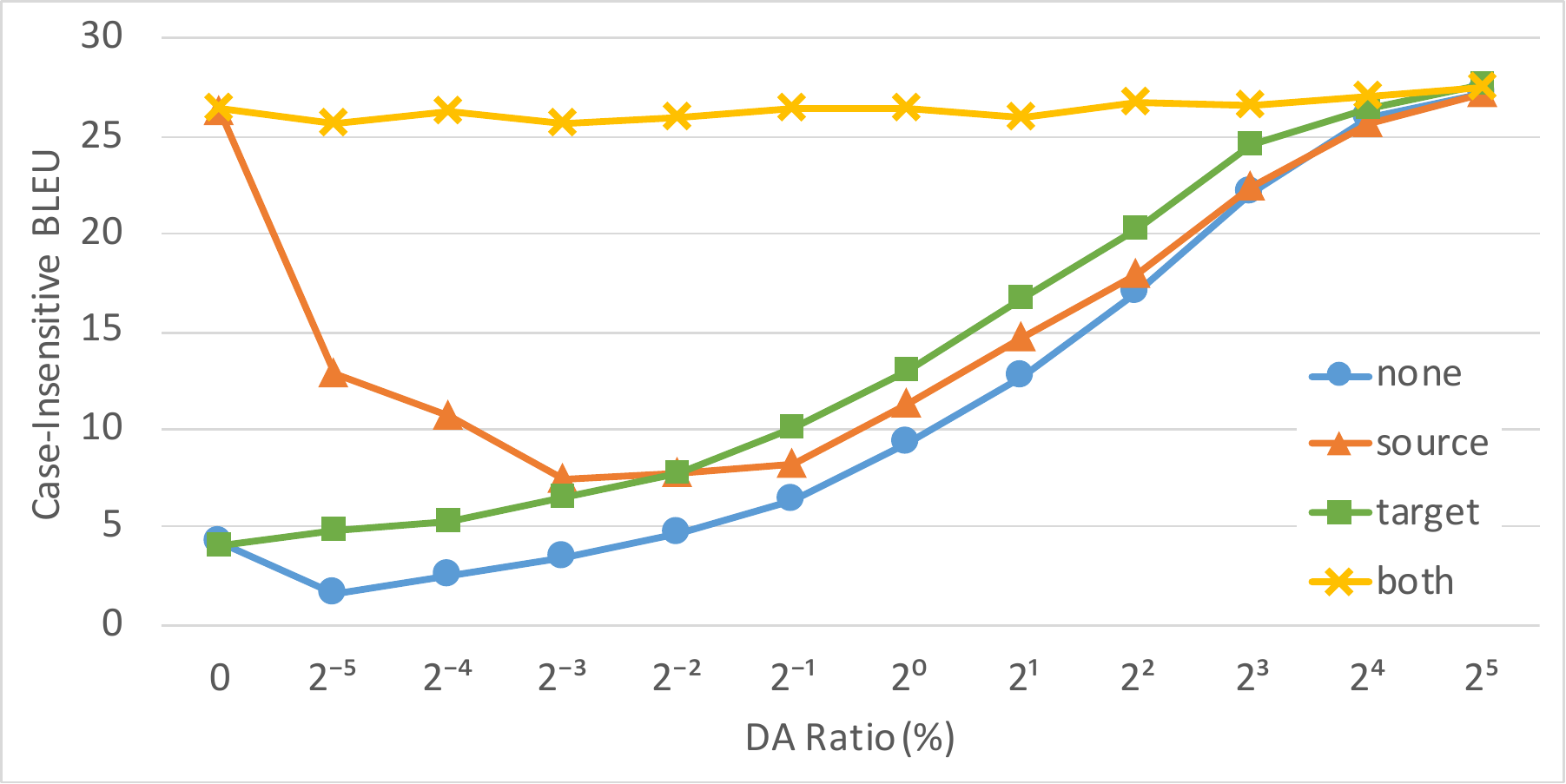}
        \caption{Using target factors (i.e., \texttt{both} and \texttt{target}) is robust to the size of augmented data in terms of case-insensitive BLEU. Otherwise (i.e., \texttt{source} and \texttt{none}), the translation quality degrades when augmenting a small amount of data.}
        \label{fig:da-bleu}
        \end{subfigure}
    \hfill
        \begin{subfigure}[b]{\columnwidth}
        \centering
        \includegraphics[width=1\textwidth]{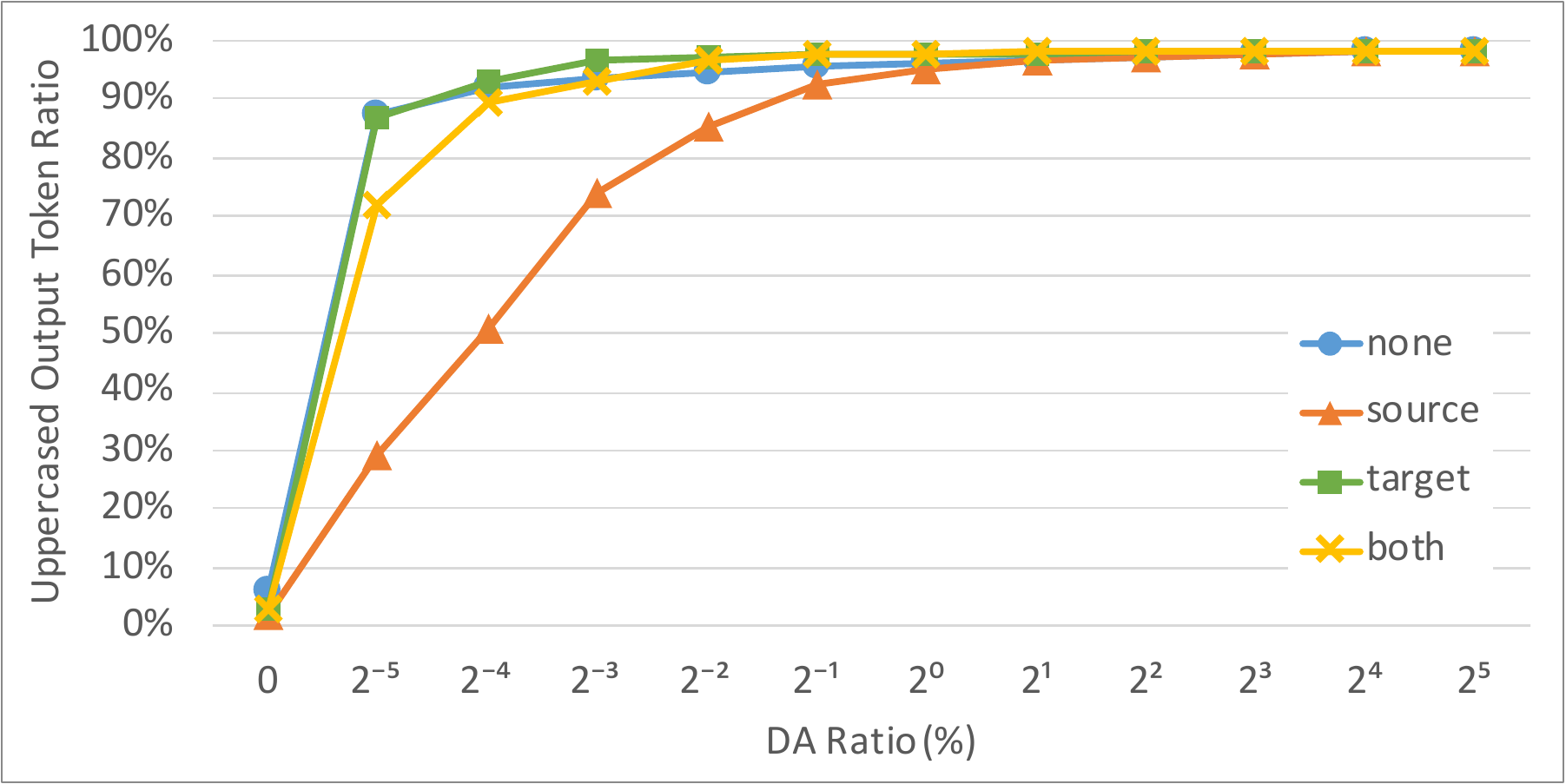}
        \caption{Using only source factors converges the slowest to preserve uppercasing to the output.}
        \label{fig:da-ratio}
        \end{subfigure}
    \hfill
        \begin{subfigure}[b]{\columnwidth}
        \centering
        \includegraphics[width=1\textwidth]{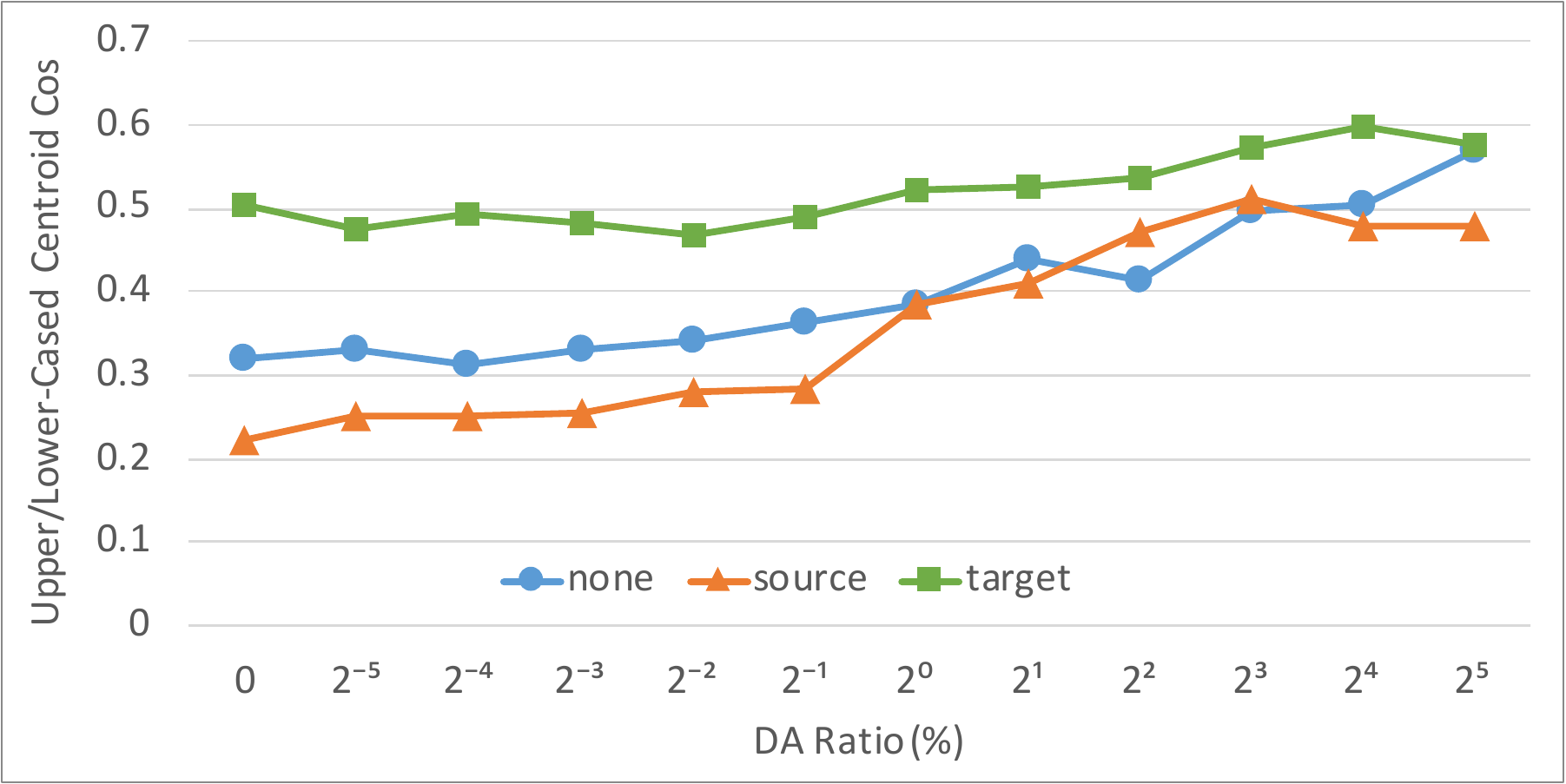}
        \caption{The increasing trend demonstrates larger subspace overlap between lowercased and uppercased embeddings.}
        \label{fig:da-embedding}
        \end{subfigure}
	\caption{Size of augmented data 
	vs.\ case-insensitive BLEU and ratio of uppercased output tokens on uppercased newstest20, as well as cosine similarity between lowercased and uppercased embeddings.}
	\label{fig:da}
\end{figure}

For a faithful model, it should be easier to alter the prediction distribution with DA. We demonstrate this on the aforementioned 5M subsampled \texttt{EN-DE} ParaCrawl data but with original casing where we vary the size of augmented data from $2^{-5}\%$ to $2^{5}\%$ of the original training data. As before, we report case-insensitive BLEU and ratio of uppercased output tokens in Figure~\ref{fig:da}.

Figure~\ref{fig:da-ratio} shows that target factor models better preserve source uppercasing when augmented with a small amount of casing-matched data, as they are more faithful to augmented data. Surprisingly, Figure~\ref{fig:da-bleu} shows (and Figure~\ref{fig:case-match-bleu} hints) that content translation quality for \texttt{source} and \texttt{none} drops significantly even with tiny (e.g., $2^{-5}\%$) augmented data. Target factors never suffer from this, no matter the initial quality (i.e., $0\%$ DA).

We hypothesize that, except when using both source and target factors, the uppercased token embeddings (which are part of the uppercased language model) are not well-trained and not well-aligned with the lowercased token embeddings before DA. 

To verify this, we separate lowercased and uppercased tokens into two groups and calculate group centroids in the embedding space. The cosine similarities between these two centroids are shown in Figure~\ref{fig:da-embedding}. The increasing trend of embedding similarities can partially explain the initial quality degradation when target factors are not used: models quickly shift to producing uppercased tokens when augmented with a small amount of data, but the uppercased target language model is under-trained, leading to low quality. 

This demonstrates that factored MT is crucial for robust DA, and using both source and target factors is optimal to achieve both good content translation and casing preservation with minimal augmentation.

\section{Conclusion and Future Work}
We investigated the simultaneous prediction of target attributes and translations focusing on two attributes: casing of words and gender of professions. We showed that this leads to translations that are more faithful to the training distribution. When applied to uppercased input, the method is robust to uncertainty in training distributions and enables data augmentation that is more stable. We leave extensions of this work to the future, including investigating joint attribute distributions and other data manipulation techniques, such as up/down-sampling.

\bibliography{anthology,custom}
\bibliographystyle{acl_natbib}

\appendix

\section{Data Pre-Processing}

We tokenize all training and development data using the Sacremoses tokenizer,\footnote{\url{https://github.com/alvations/sacremoses}} except the French data, which is tokenized alongside with the morphological annotation by spaCy.\footnote{\url{https://spacy.io/}} Words are segmented using byte pair encoding (BPE, \citealp{sennrich-etal-2016-neural}) with 32K operations. Source and target subwords share the same vocabulary.

For experiments involving casing attributes, we true-case the training data to reduce the vocabulary before training BPE models but recover the original cases after applying BPE segmentation. The true-case of a word is simply its most frequent case variation.

\section{Factor Deduction}

We consider four case factors, namely uppercased, capitalized, lowercased, and undefined (e.g., punctuations). Factors are deduced at the subword level, so we prohibit BPE merge operations that lose case information, such as \texttt{Wi@@} and \texttt{Fi} are not merged no matter how frequent \texttt{WiFi} is.

We consider three gender factors annotated by spaCy, namely masculine, feminine, and unknown. Gender is associated with each word, so we broadcast the annotation to all its subwords.

\section{NMT Model Configuration}

We build NMT models with the Transformer-base architecture \citep{vaswani-etal-2017-attention} except we use 20 encoder layers and 2 decoder layers as recommended by \citet{domhan-etal-2020-sockeye}. The source embeddings, target embeddings, and the output layer's weight matrix are tied \citep{press-wolf-2017-using}. Training is done on 8 GPUs with Sockeye~2's large batch training. It has an effective batch size of 262,144 tokens, a learning rate of 0.00113 with 2000 warmup steps and a reduce rate of 0.9, a checkpoint interval of 125 steps, and learning rate reduction after 8 checkpoints without improvement. After an extended plateau of 60 checkpoints, the 8 checkpoints with the lowest validation perplexity are averaged to produce the final model parameters.

\section{Generating German Nouns}

We have demonstrated that using both source and target factors is the best option in the application of translating uppercased input with data augmentation.

One may argue that this model could just learn to copy all source factors to the target. We examine this by testing the generated casing of nouns in German. In German, nouns are always capitalized while it is not the case in English, so the model needs to predict nouns' casing based on some information beyond the source casing. We extracted 4,403 test pairs from newstest2011 to newstest2020, where the German side has 4 times more capitalized tokens than the English side. Results show that all 48 models we investigated (4 factor configurations $\times$ 12 augmented data sizes) yield almost the same case-sensitive BLEU ($26.76\pm0.15$) and ratio of capitalized tokens ($30.38\pm0.04\%$) on this subset. It indicates that all methods, including \texttt{both}, learn to predict German nouns’ casing well, and it is not hurt by augmenting only uppercased data.

\section{Human Translation of Professions}
\label{appendix:professions}

Combining three profession lists collected by \citet{rudinger-etal-2018-gender}, \citet{zhao-etal-2018-gender}, and \citet{prates-etal-2020-assessing} yields 1054 distinct English gender-neutral professions. Professions can be either single words (e.g., \texttt{singer}) or phrases (e.g., \texttt{flight attendant}). We request both masculine and feminine French translations from paid professional translators. The instruction to translators contains examples: ``If the profession is \texttt{driver} we would like both \texttt{conducteur} and \texttt{conductrice}. Ideally we want equivalent masculine and feminine words if such equivalents exist (so not \texttt{chauffeur} and \texttt{conductrice})." All translations were done by a single translator and QA'd by a second translator. The translations are available at \url{https://github.com/amazon-research/gendered-profession-translations}.

The translators returns 861 gender-distinct translation pairs, meaning masculine and feminine translations are not the same. Then we count phrases in each pair and exclude pairs where the total phrase count is less than five in the training data (i.e., $\mathrm{count}(feminine)+\mathrm{count}(masculine) < 5$). 355 pairs remain after this filtering. Finally, we exclude two outlier professions: (1) \texttt{cook}, as its feminine translation \texttt{cuisini\`ere} commonly refers to cooking equipment; (2) \texttt{flight attendant}, as its feminine translation \texttt{h\^otesse de l'air} is longer than its masculine translation \texttt{steward}, which leads to biased translation score. In addition, \texttt{steward}, the masculine translation of \texttt{flight attendant}, works for both masculine and feminine. This results in 353 pairs used in our experiment.

\section{In-Context Gender Translation}
\label{appendix:gender}

\begin{table}[t]
\centering
\begin{tabular}{l|r|r|r|r}
    & \multicolumn{4}{c}{WinoMT ($F_1$/Accuracy)} \\
    & masc & fem & pro & anti \\
    \hline
    baseline & 62.2 & 53.3 & 70.6 & 44.7 \\
    target factors & \bf 64.4 & \bf 61.3 & \bf 77.3 & \bf 48.2
\end{tabular}
\caption{Baseline and target factor models are evaluated on WinoMT. Target factors improve across all four WinoMT subsets: \textbf{masc}uline $F_1$, \textbf{fem}inine $F_1$, \textbf{pro}-stereotypical accuracy, and \textbf{anti}-stereotypical accuracy.}
\label{tab:gender}
\end{table}

We also evaluate \texttt{EN-FR} models on WinoMT \citep{stanovsky-etal-2019-evaluating} to assess the translation accuracy of professions in context. This data set contains examples of professions for which the gender is disambiguated by the context, such as \textit{mover} in \textit{The secretary asked the mover when she is available}. Results show that simultaneously predicting gender markers improves gender translation across all subsets (Table~\ref{tab:gender}). We hypothesize that the context-conditioned gender distribution in the training data is better preserved, because the training loss of gender prediction additionally encourages the model to attend more to the context, hence disambiguate the gender better. We leave the verification as future work.

\end{document}